\documentclass[10pt]{article}
\usepackage{indentfirst}
\usepackage{amssymb}
\usepackage{amsmath}
\usepackage[dvipsnames]{xcolor}
\usepackage{color}
\usepackage{graphicx}

\begin{document}
\title{A Delay-free Control Method Based On Function Approximation And Broadcast For Robotic Surface And Multiactuator Systems} 
\author{Yuchen Zhao}
\date{}
\maketitle

\let\thefootnote\relax
\footnotetext{Yuchen Zhao is with School of Automation, Southeast University, Nanjing 210096, Jiangsu, China, and also with Ministry of Education Key Laboratory of Measurement and Control of Complex Systems of Engineering, Southeast University, Nanjing 210096, Jiangsu, China. yuchen.zhao078@seu.edu.cn\\
*This work was supported by the start-up research fund of Southeast University.
This work has been submitted to the IEEE for possible publication. Copyright may be transferred without notice, after which this version may no longer be accessible.} 

\begin{abstract} 
Robotic surface consisting of many actuators can change shape to perform tasks, such as facilitating human-machine interactions and transporting objects.
Increasing the number of actuators can enhance the robot's capacity, but controlling them requires communication bandwidth to increase equally in order to avoid time delays.
We propose a novel control method that has constant time delays no matter how many actuators are in the robot.
Having a distributed nature, the method first approximates target shapes, then broadcasts the approximation coefficients to the actuators, and relies on themselves to compute the inputs.
We build a robotic pin array and measure the time delay as a function of the number of actuators to confirm the system size-independent scaling behavior.
The shape-changing ability is achieved based on function approximation algorithms, i.e. discrete cosine transform or matching pursuit.
We perform experiments to approximate target shapes and make quantitative comparison with those obtained from standard sequential control method.
A good agreement between the experiments and theoretical predictions is achieved, and our method is more efficient in the sense that it requires less control messages to generate shapes with the same accuracy.
Our method is also capable of dynamic tasks such as object manipulation.

\smallskip
\textbf{key words}: Cellular and Modular Robots, Distributed Robot Systems, Factory Automation, Haptics and Haptic Interfaces, Virtual Reality and Interfaces

\end{abstract}

\section{INTRODUCTION}
Recent developments in robotics and human-machine interface lead to various advances in robotic surface~\cite{walker2017continuum}.
The robot typically consists of many independent actuation modules arranged in an array and can serve as a shape display or refreshable braille~\cite{leithinger2014physical,chen2023novel}, haptic interface~\cite{nakagaki2016materiable,abtahi2018visuo-haptic,nakagaki2019inforce}, conveyor~\cite{uriarte2019control,chen2024trajectory}, adaptive structures~\cite{Wang2019design,salerno2020ori-pixel}, molding tools~\cite{tian2022soft, adapa_adaptive_moulds2024adaptive}, or treadmill floor~\cite{smoot2019floor}.
Developments in soft robotics also bring new designs and solutions to meet its demand of many actuators~\cite{johnson2023multifunctional,robertson2019compact,liu2021robotic}.

The capability of robotic surface is related to its number of actuators.
Adding more actuators can lead to high-resolution patterns for complex expressions and tasks.
However, controlling many actuators is challenging, as generating control inputs for them would require a large amount of resources such as physical space, equipment and communication bandwidth~\cite{winck2013dimension}.
A noticeable quantity is the time delay $\tau$ between the time when a control message is sent to the first actuator and the time when the last actuator reaches its target position. 
A small $\tau$ implies fast refresh rate, which is preferable in real-time tasks.
$\tau$ mainly consists of: (1) the communication time during which a central computer sends control messages to each actuator; (2) actuator dynamics.
Standard communication methods send control messages in a sequential fashion~\cite{follmer2013inform,xue2023arraybot,siu2018shapeshift,stanley2016closed-loop,leithinger2010relief}, so $\tau$ is proportional to the number of actuators, i.e.\ the system size $N$.
To control more actuators, strategies such as using multiple communication channels, sharing one motion controller in a small group of motors, or perform multithreading in the central computer have been employed~\cite{johnson2023multifunctional,follmer2013inform,xue2023arraybot,siu2018shapeshift}.
These methods may result in cumbersome software or hardware designs, and are not scalable when $N$ increases.

A more scalable approach is to drive each row and column of the actuator array similar to the matrix drive technique used in LED displays~\cite{chen2011handbook}.
In a series of works, Zhu, Winck and Book et~al.\ show that the scaling of $\tau$ can be reduced to $\sqrt{N}$.
They developed a control loop structure based on singular value decomposition that can drive the $2\sqrt{N}$ rows and columns of a hydraulic cylinder robotic array to any shape~\cite{zhu2004practical, zhu2006construction,winck2012control, winck2012control-1, winck2012command, winck2013dimension, winck2015svd, winck2016passivity}.
In this method, a `control coupler' valve is also needed for each cylinder to integrate the row and column control messages, which are fluid pressures~\cite{ferguson2020multiplicative}.
More recently, Jadhav et al.\ designed a compact fluidic logic module to regulate the input row and column pressures for a pneumatic soft linear actuator array~\cite{jadhav2023scalable}.
Besides fluidic actuators, a robotic surface made of ionic polymer stripes are controlled using peripheral voltages based on pre-trained neural networks~\cite{Wang2023passively}.
While reducing time delay, this method compromises shape accuracy due to low-rank approximations, and still lack scalability because it has a system size-dependent time delay.
   \begin{figure}[thpb]
      \centering
      \includegraphics[width=3.1in]{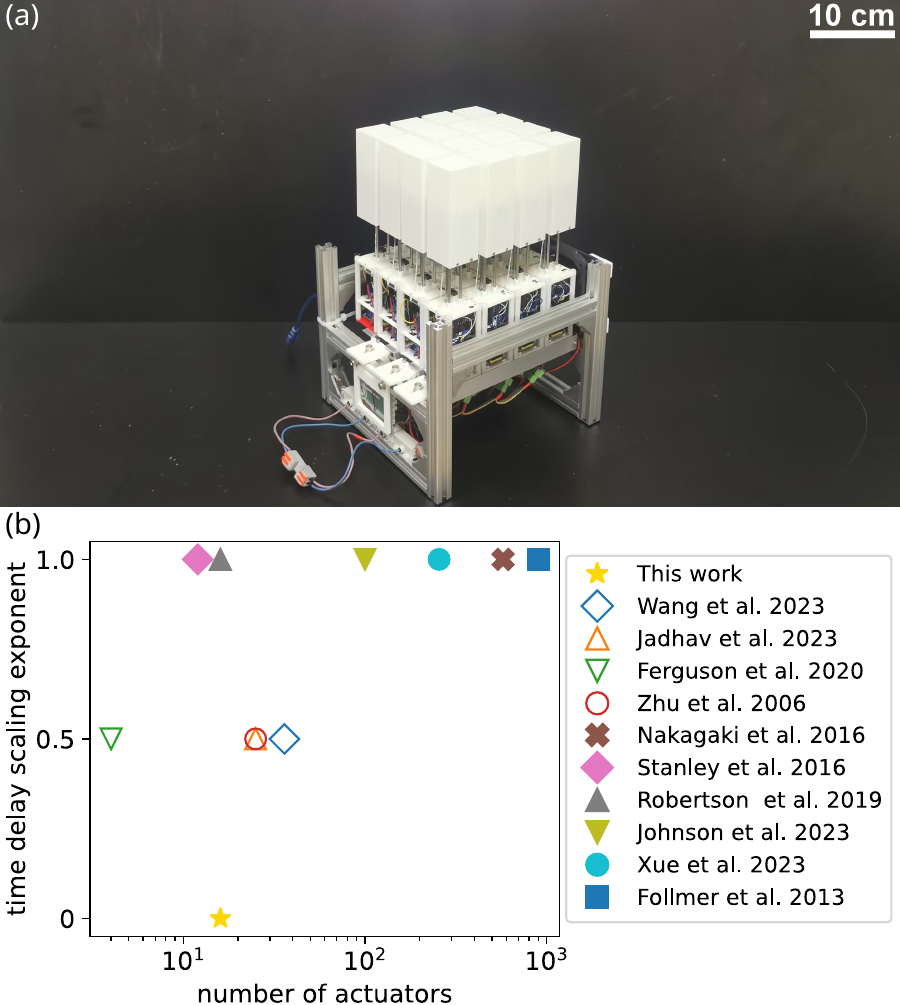}
      \caption{
      (a) The 4$\times$4 pin array; (b) time delay scaling exponent $\alpha$ is plotted vs.\ the number of actuators. 
      Our work is highlighted in a yellow star.}
      \label{fig:overall}
   \end{figure}

This paper proposes a new control method for robotic surface that has system size-independent time delay.
The central computer broadcasts features of the target shape to individual actuation modules and has them calculate their inputs on-site.
The motivations are: (1) neighboring actuators usually have similar control inputs, which could be approximated using interpolation and hence no need to send the inputs to every one of them; (2) complex patterns may be simply parametrized, such as the Gaussian function pattern used in object manipulation tasks~\cite{johnson2023multifunctional} in which only two center coordinates are important.
These changes in control method result in a size-independent scaling of time delay $\tau \propto N^{\alpha},\,\alpha=0$.
We test this new method on a $4\times 4$ pin array setup, as shown in Fig.~\ref{fig:overall}(a).
$\alpha$ in existing works are also summarized and plotted vs.\ system size in Fig.~\ref{fig:overall}(b).
We experimentally validate the time delay scaling and compare it to the sequential control method.
In order to achieve any shape, we use function forms with universal approximation property and employ discrete cosine transform (DCT) and matching pursuit (MP) algorithm~\cite{mallat1993matching} to compute the coefficients as pattern features.
We further characterize shape change capability by displaying 6 distinct shapes and measure their height profiles.
We also perform object manipulation experiments to demonstrate the simplicity in generating shapes for moving objects.
Contributions of this work are summarized as below:
\begin{itemize}
\item A new control method for robotic surface that can achieve a system-size independent time delay.
\item An implementation of this method in a $4 \times 4$ pin array, and we use two algorithms, discrete cosine transforms or matching pursuit, to compute the shape approximation coefficients. 
\item Experimental validation the time-delay scaling with and without the presence of actuator dynamics.
\item Systematic experiments that characterize the shape-changing capacity, and success in dynamic object manipulation tasks.
\end{itemize}

The rest of the paper is organized as follows.
In Sec.~\ref{sec:control}, we present the control method, the algorithms and functions with universal approximation properties to compute the coefficients.
In Sec.~\ref{sec:setup}, we describe the mechanics and electronics of the pin array.
In Sec.~\ref{sec:expt}, we present our experimental validation of the time-delay scaling, quantification of shape-changing capacity, and object manipulation tasks.
Sec.~\ref{sec:conclusion} contains a discussion and concluding remarks.

\section{APPROXIMATION AND BROADCAST}\label{sec:control}
   \begin{figure}[thpb]
      \centering
      \includegraphics[width=3.1in]{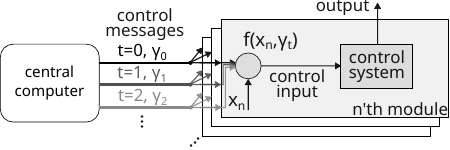}
      \caption{
      An illustration of our control method. At time $t$, the control message $\gamma_t$ is broadcast to all modules. In the n'th module, the control input is calculated via a function $f$ and its arguments $\gamma_t$ and $x_n$.}
      \label{fig:controlSchematics}
   \end{figure}
In Fig.~\ref{fig:controlSchematics}, we illustrate the information flow in our control method.
Instead of sending control messages to each actuator sequentially as in the standard method, we broadcast control messages $\gamma_t$ to all actuation modules at each time step $t$.
In our case, an actuation module refers to an independent system that includes an actuator, a sensor, and microcontrollers (Sec.~\ref{sec:setup}). 
$\gamma_t$ does not explicitly contain references or control inputs to the module.
It is a set of coefficients used by each module to compute the actual reference or control input.
Although the modules receive the same control messages, they can use a single function form $f(x_n,\gamma_t)$ with universal approximation properties and a local parameter $x_n$ to compute their inputs, making it possible for the robot to approximate any target surface profile.
Note that $x_n$ is an identification value stored on the n'th module for computing the input, and it must be different for each module.

Based on the above intuition, we choose two such functions to test the time delay scaling and shape-changing ability.
The cosine function is used because of an analogy between shape display and the JPG image compression, the latter of which uses discrete cosine transform (DCT).
The input $f_n$ to the actuator on n'th module is represented as:
\begin{equation}
f_n = a_0 + 2\sum_{t=1}^{N-1} a_t cos(k_t(2x_n+1)),\quad k_t=\frac{\pi t}{2N}
\label{eqn:dct}
\end{equation}
\noindent where $a_t$ and $k_t$ are the amplitude and wave vector at time $t$, and $\gamma_t\equiv(a_t,k_t)$.
Technically, we compute DCT using the Scipy package \texttt{fftpack} on the central computer.
To construct an accurate target shape, several $\gamma_t$ are broadcast to all modules.
Upon receiving the messages, each module compute the input $f_n$ according to (\ref{eqn:dct}) with their own $x_n$.
There is no system size-dependent time delay since all modules simultaneously receive $\gamma_t$ and compute $f_n$ in parallel.
So we have:
\begin{equation}
\tau \propto N^{\alpha} ,\quad \alpha=0
\label{eqn:timescaling}
\end{equation}

A drawback of (\ref{eqn:dct}) is that it is not efficient in representing spatially localized patterns.
Therefore, we use time-frequency functions paired with the matching pursuit algorithm (MP)~\cite{mallat1993matching,mallat2008wavelet} to capture both extended and localized patterns:
\begin{equation}
f_n = \sum_{t=0}^{\infty} a_t g_{s_t}(x_n-p_t) \cos(\frac{2\pi k_t}{N}x_n+\phi_t)
\label{eqn:mp}
\end{equation}
\noindent where $g_s(x)=\sum_{j=-\infty}^{+\infty}\exp(-\pi (\frac{x-jN}{s})^2)$ is a Gaussian function made periodically on the domain $[0,N]$.
This formula is a discrete implementation of matching pursuit for a dictionary of time-frequency atoms~\cite{mallat1993matching}.
The coefficients in $\gamma_t\equiv(a_t, s_t, p_t, k_t, \phi_t)$ are the amplitude, scale, position, frequency, and phase of the time-frequency atom, respectively.
They control the span and concentration of the pattern in real and frequency space, giving more flexibility in capturing the features in the target shape.
We implement the MP algorithm described in \cite{mallat1993matching} in Python.
In short, it is a sequential procedure in which we minimize the norm of the residual shape after subtracting each projection of a time-frequency atom onto the current shape.
This minimization is achieved in two steps, a matching step where the best atom is selected from a redundant dictionary and a pursuit step where the coefficients are fine-tuned to maximize the projection.
The algorithm converges exponentially as proven in \cite{mallat1993matching}, and in practice we find only a few terms are needed to approximate a pattern with good accuracy (see Sec.~\ref{sec:shapeTest}).

For object manipulation tasks, we use Gaussian radial basis function (RBF), which has been used in \cite{johnson2023multifunctional}:
\begin{equation}
f_n = \sum_{t=0}^{\infty} a_t \exp(- \frac{(x_n-d^{(x)}_t)^2+(y_n-d^{(y)}_t)^2}{\sigma_t^2} )
\label{eqn:gs}
\end{equation}
\noindent where the coefficient in $\gamma_t\equiv(a_t,\sigma_t,d^{(x)}_t,d^{(y)}_t)$ are the amplitude, width, and center coordinates of each Gaussian function.
$(x_n,y_n)$ are two identification values stored on the n'th module, representing the 2D physical coordinates of the module in the array.
Although Gaussian RBF is capable of universal approximation~\cite{park1991universal}, for the object manipulation tasks we only use one term in (\ref{eqn:gs}) to hold the object.

\section{ACTUATION MODULE AND ROBOT}\label{sec:setup}
\subsection{Linear Actuator Design}
\begin{figure}[thpb]
      \centering
      \includegraphics[width=3.1in]{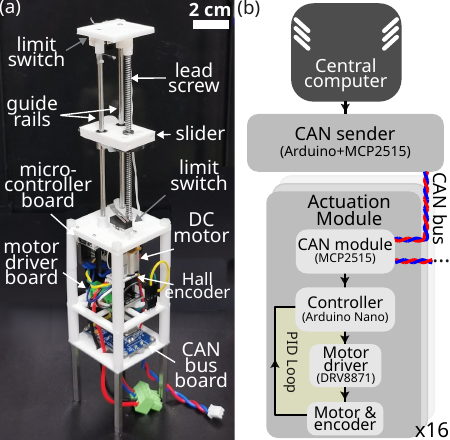}
      \caption{
      (a) A picture of a single linear actuation module. The rectangular cover is removed to expose mechanical components; (b) block diagram of the electronic system. The arrows indicate information flows.}
      \label{fig:setup}
   \end{figure}

The pin array robot has a modular design and consists of 16 identically built linear actuation module arranged in an $85$\,mm-long square area (Fig.\ref{fig:overall}(a)).
As shown in Fig.\ref{fig:setup}(a), the module is about $47$\,mm wide and $200$\,mm long.
A lead screw of $100$\,mm in length and $2.5$\,mm in pitch converts rotary motion of a DC motor to linear motion.
The screw is attached to a slider, with two additional guide rails parallel to the lead screw to reduce friction.
Two limit switches are installed at the two ends to prevent overshoot that may damage the motor, and the overall arrangement of mechanical parts results in a linear stroke of $70$\,mm. 
A complete module also has a rectangular cover attached to the slider (see Fig.~\ref{fig:overall}(a)).
The DC motor (Tianqu Motor, N20VA, 1:10) has a rated full speed of $50$ revolution per second, leading to a nominal speed of $125$\,mm/s of the linear motion.
The motor's tail has a Hall rotary encoder to measure the angular position of the shaft, defined as $e_n$ for the n'th module.
The vertical position $h_n$ of the linear actuator is proportional to $e_n$.
All mechanical components and electronics are mounted on 3D printed frames, and the modules are mounted on a portable aluminum frame.
When varying the system size $N$, we simply connect or disconnect modules from the robot.

\subsection{Position Control and Broadcast Communication}
The electronics block diagram is shown in Fig.~\ref{fig:setup}(b).
The controller of the actuator is an Arduino Nano board, which is programmed as a closed-loop control system for shaft position $e_n$ and can receive control messages from a central computer.
A PID controller running every $16$\,ms ($62.5$\,Hz) is in use.
Two additional modules are connected to the controller board.
A DRV8871 board is used to drive the motor with standard pulse width modulation (PWM) technique.
An MCP2515 CAN bus board is used for receiving control messages.
All actuation modules are on one CAN bus.
CAN bus is chosen because we can perform both sequential and broadcast communication with little modification in the software.
While identical in hardware, there is the unique identification variable $x_n$ ranging from $0$ to $15$ in the n'th module's software.
This variable is involved in control input calculation in (\ref{eqn:dct}), (\ref{eqn:mp}), and (\ref{eqn:gs}), or acts as the CAN message identifier in the sequential control method.
A central computer (Raspberry Pi 4B) is used to generate target shapes $\{f_n\}$ and the control messages $\{\gamma_t\}$ to drive the modules.
In our method, we broadcast a single $\gamma_t$ in one standard CAN data frame (which holds 8 bytes of data), meaning the coefficients are represented with limited resolution.
The data frame also contains information on which function form to use, so we can switch between different approximation methods.
An additional Arduino Nano and an MCP2515 module serve as the CAN bus sender that interfaces with the central computer and the modules.

\section{EXPERIMENTS}\label{sec:expt}
\subsection{Time Delay Scaling}\label{sec:scalingTest}
\begin{figure}[thpb]
      \centering
      \includegraphics[width=3.1in]{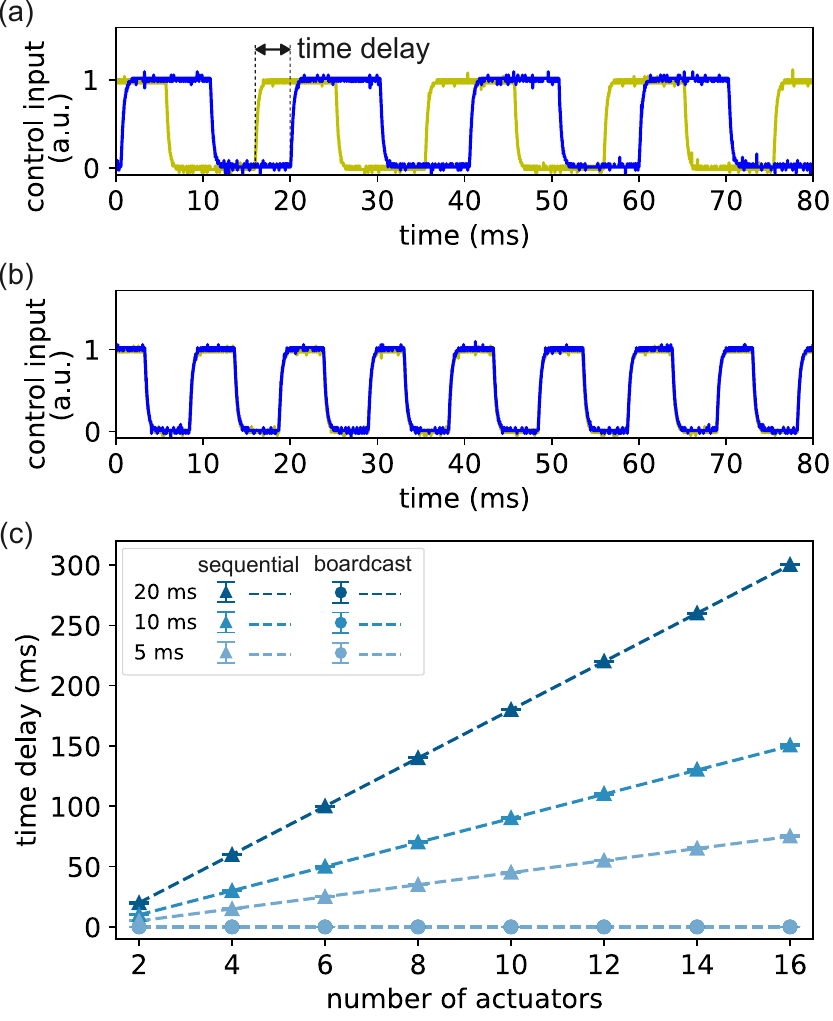}
      \caption{
      Experimental time delay scaling without actuator dynamics. (a) The control inputs of two actuators when using sequential control method. The yellow (blue) line is the first (last) actuator in a two-actuator system; (b) the control inputs of the same actuators when using our control method;
      (c) averaged time delay is plotted as a function of the number of actuators for the two control methods and at different communication rates (expressed in $T_{\rm msg}$). Each point is an average of at least $20$ time delay values observed in (a) or (b). Darker color represents larger $T_{\rm msg}$. The triangles (circles) are experimentally measured time delays with the sequential (our) control method, and the dashed lines are theoretical predictions.}
      \label{fig:scaling-binary}
   \end{figure}

We first perform an experiment to measure the time delay when only communication delay is present.
The central computer refreshes the robot between two uniform patterns, $\{ f_n=0 \}_{n=1}^N$ and $\{ f_n=1 \}_{n=1}^N$, at a constant rate using the sequential or our method in (\ref{eqn:dct}).
To measure the computed control input $f_n$ on each actuator, we correlate the variable with a PWM output on the controller board, and convert the PWM output using a digital-to-analog converter based on an LCR low-pass circuit.
The analog signals from the first and last module in the robot are simultaneously measured on an oscilloscope (Rigol DS1202EZ).
Typical signals are shown in Fig.~\ref{fig:scaling-binary}(a) and (b) for a system with $N=2$ actuators.
We extract the time delay using normalized cross-correlation and average over at least $20$ values.

The averaged time delay is plotted as a function of $N$ for both control methods in Fig.~\ref{fig:scaling-binary}(c).
As expected, for the sequential control method, the time delay shows a linear increasing trend, and the slope equals to the time period $T_{\rm msg}$ to send one control message.
We vary $T_{\rm msg}$ from $5$\,ms to $20$\,ms, and the predicted trends agree well with experimental results. 
In contrast, for our method, the time delay remains at zero for all $N$ and all $T_{\rm msg}$, showing that there is no time delay due to system size effect or communication rate.
Therefore, we confirm that our method indeed achieves a system size-independent time delay when no actuator dynamics is present.

In the second experiment, we take the actuator dynamics into account by measuring the time delay between shaft position $e_n$ of the first and the last actuator.
We send a traveling wave pattern to the robot, in which each actuator is driven to different time-varying positions.
The traveling wave is a quarter of a moving sinusoid over $N$ actuators, given by:
\begin{equation}
f_n = \sin (k_N x_n - v t)
\label{eqn:traveling-wave}
\end{equation}
\noindent where $k_N=\frac{\pi}{2(N-1)}$ is the system-size dependent wave vector chosen such that the last actuator $x_N = N-1$ always has a quarter phase, $k_N x_N = \frac{\pi}{2}$.
$v$ is the traveling speed, and $t$ is the current time.
The control message is sent every $T_{\rm msg} = 5$\,ms.
For our method, we express $f_n$ using a simplified version of (\ref{eqn:mp}) because (\ref{eqn:traveling-wave}) can be rewritten as $f_n = a\cos (k_N x_n - \pi/2) + b\cos (k_N x_n)$, where $a=\cos (v t)$ and $b=-\sin (v t)$.
The coefficients $k_N$, $a$, and $b$ are broadcast to all modules in one CAN data frame.

\begin{figure}[thpb]
      \centering
      \includegraphics[width=3.1in]{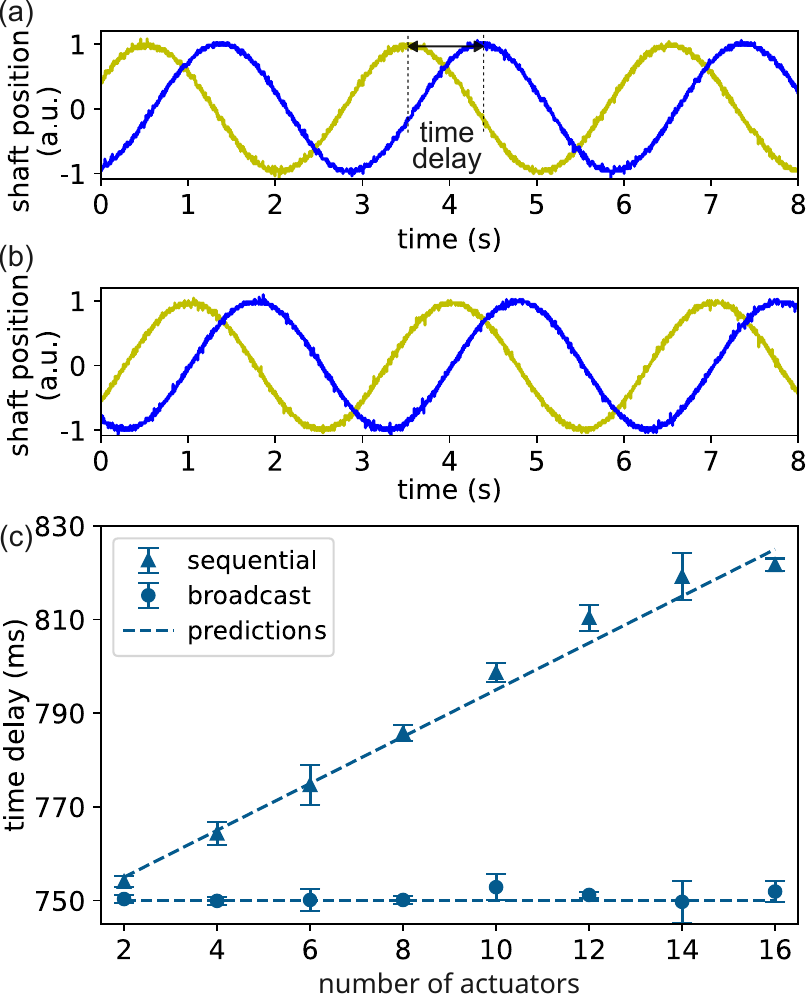}
      \caption{
      Experimental time delay scaling with actuator dynamics. (a) The motor shaft angular position when using the sequential control method. The yellow (blue) trace is the first (last) actuator in a 16-actuator system, with their time delay indicated in dashed line; (b) the shaft position of the same actuators when using our control method; (c) time delay is plotted as a function of the number of actuators for the two control methods. The triangles and circles are experimentally measured time delays with the sequential and our control method, respectively. Each point is an average of at least 6 time delays, and the errorbar is one standard deviation. The dashed lines are theoretical predictions.}
      \label{fig:scaling-traveling}
   \end{figure}

The observed traveling waves for both methods are shown in Fig.~\ref{fig:scaling-traveling}(a) and (b), and the average time delay $\tau$ as a function of the number of actuators $N$ are shown in Fig.~\ref{fig:scaling-traveling}(c).
For our method, $\tau$ stays constant for all $N$, hence validating our claim when both signal transmission and actuator dynamics are present.
According to (\ref{eqn:traveling-wave}), the time delay between the first and last actuator should be the time to catch up their phase difference, i.e. $\tau = \frac{k_N(x_N-x_1)}{v} = \frac{\pi}{2v}$.
In all experiments, we set $v=\frac{2\pi}{T}$ and $T=3000$\,ms.
So $\tau = T/4 = 750$\,ms, which agrees with the experiments in Fig.~\ref{fig:scaling-traveling}(c).
For the sequential control method, $\tau$ is larger than the theoretical $750$\,ms and linearly increases as $N$ increases.
When one module is added to the system, it takes an additional amount of time to transmit control messages to that module, so the time delay for the sequential control method is $\tau_{\rm seq} = T/4 + T_{\rm msg}(N-1)$.
This prediction is plotted as a dashed line in Fig.~\ref{fig:scaling-traveling}(c), which also agrees with the experiments.

\subsection{Characterization of Shape Change}\label{sec:shapeTest}
\begin{figure}[thpb]
      \centering
      \includegraphics[width=3.1in]{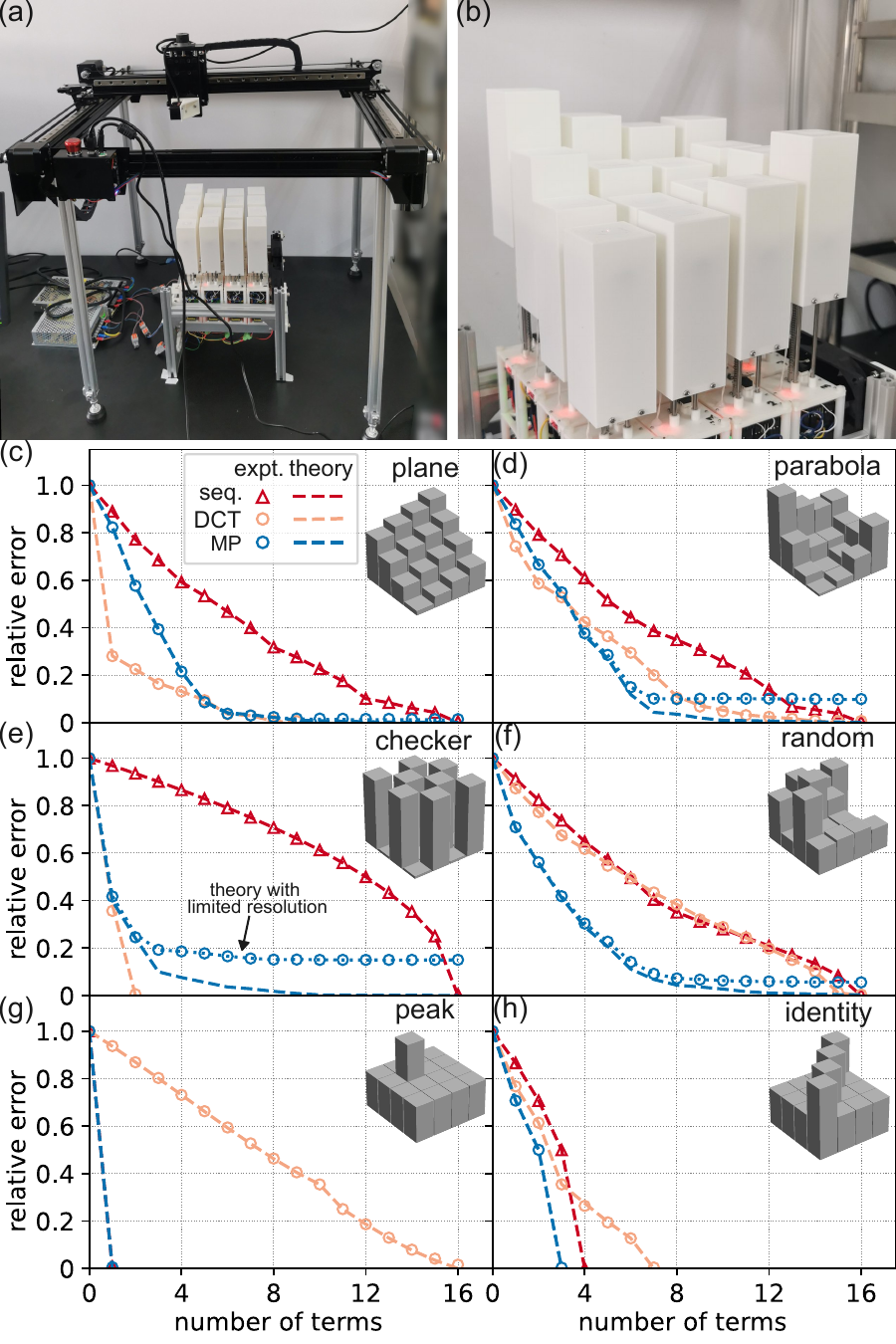}
      \caption{
      Characterization of shape-changing ability. (a) The experimental setup for shape measurement; (b) the parabola shape displayed by the robot; (c)-(h) relative error of shape is plotted as a function of number of terms used to approximate the shape. The triangles (circles) are from the sequential (our) control method. The orange and blue colors of the circles correspond to DCT and MP algorithm, respectively. Each data point is an average of 6 independent runs, and the errorbar is smaller than the marker size, hence not shown. The dashed lines are theoretical predictions calculated with 64-bit float point numbers, and the dotted blue lines are calculated with 16 and 8-bit resolution-limited numbers. The inset displays the target shape.
      }
      \label{fig:shapeTest}
   \end{figure}

	\begin{table}[h]
	\caption{Name and expression of the $6$ shapes.}
	\label{tab:shape}
	\begin{center}
	\begin{tabular}{|c|c||c|c|}
	\hline
	\textbf{Name} & \textbf{Expression} & \textbf{Name} & \textbf{Expression}\\
	\hline
	identity & a 4$\times$4 identity matrix & plane & $z=x+2y$\\
	\hline
	parabola & $z=2 x^2+ 3 y^2- 3 xy$ & checkers & checkerboard pattern\\
	\hline
	peak & \begin{tabular}[c]{@{}c@{}}raise a single module\\at $(x=1,y=2)$\end{tabular} & random &\begin{tabular}[c]{@{}c@{}}random uniform\\distribution\end{tabular}\\
	\hline
	\end{tabular}
	\end{center}
	\end{table}
We drive the robot to $6$ shapes and quantify the errors between target and measured shapes.
The shape measurement apparatus is shown in Fig.~\ref{fig:shapeTest}(a).
A laser distance meter (Shanghai Kedi, KG01) is used to measure the height change of each actuator, and the scan process is automated via a homemade Cartesian robot.
The system has an accuracy of $0.2$\,mm.
The 6 shapes are listed in Table~\ref{tab:shape}.
All shapes are represented as a $4\times 4$ matrix, where the elements are actuator vertical positions $h_n$.
The actuation module's planar coordinates $(x_n,y_n)$ are taken from $\{0,1,2,3\}$, and we stretch the vertical scale so that each shape can fill the entire $70$\,mm stroke of the actuator.
The parabola shape displayed by the robot is shown in Fig.~\ref{fig:shapeTest}(b), and all shapes are shown as the inset in Fig.~\ref{fig:shapeTest}(c) to (h).
For each run of the experiment, we start with the robot leveled at half stroke length.
Then the robot is actuated through a series of intermediate shapes.
Following each shape actuation, we perform a height scan to track the shape change, and compute the relative error between the intermediate shape and the target shape using root squared error. 
The intermediate shape is achieved incrementally with additional information from one control message $\gamma_t$ that contains all coefficients in one approximation term in (\ref{eqn:dct}) and (\ref{eqn:mp}).
For the sequential control method, the term is the just reference vertical position $h_n$ of a single actuator.
For our method, we test both function approximation formula in (\ref{eqn:dct}) and (\ref{eqn:mp}).
To compute the coefficients, we reshape the $4\times 4$ shape matrix into a vector and apply DCT or matching pursuit.
On determining the order of shape actuation, we choose the term with a higher amplitude or $h_n$ first. 

The relative error as a function of the number of terms (or equivalently the number of control messages) is shown in Fig.~\ref{fig:shapeTest}(c)-(h).
Both algorithms of our control method outperform the sequential control method in the sense that they require less terms to approximate the target shape for the same error.
For example, it takes $11$ terms for the sequential method to approximate the parabola shape to a $20\%$ relative error, while it takes $6$ and $7$ terms for the MP and DCT, respectively (Fig.~\ref{fig:shapeTest}(d)).
We also find that within our methods, MP can approximate both extended and localized patterns with less terms, while DCT approximates extended patterns better than localized patterns.
For example, for the peak shape, it takes only one term for MP to reach a perfect match, while DCT needs all $16$ terms.
The sequential method also needs one term, because it only needs to actuate a single module in this case.
The relative errors obtained from experiments are well captured by theoretical predictions based on (\ref{eqn:dct}) and (\ref{eqn:mp}), except in the parabola, checker, and random cases, MP leads to residual errors that are not captured by (\ref{eqn:mp}).
These residual errors are due to limited coefficient resolution used in computing the approximation terms.
To include the five coefficients from (\ref{eqn:mp}) into one CAN data frame, $s_i$ and $a_i$ have 16-bit resolution.
$p_i$, $k_i$, and $\phi_i$ have 8-bit resolution.
By recalculating intermediate shapes using these less precise numbers, a good agreement with the experiments can be achieved, as shown by the dotted lines in Fig.~\ref{fig:shapeTest}(d)-(f).
\subsection{Object Manipulation}\label{sec:manipulation}
\begin{figure}[thpb]
      \centering
      \includegraphics[width=3.1in]{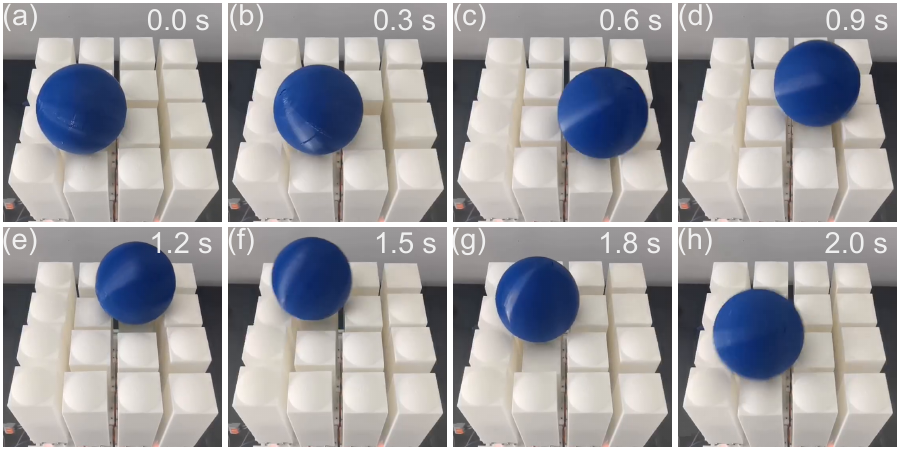}
      \caption{
      Illustration of object manipulation capability. Figure (a) to (h) are snapshots of a blue sphere following a rectangular trajectory driven by the robot. There is a spherical cap attached to the top of each rectangular cover to prevent the object stabilizing itself on the top.}
      \label{fig:manipulation}
   \end{figure}

We demonstrate object manipulation capability of the robot based on our method.
The object is a 3D printed sphere of diameter $80$\,mm and weighs $25$\,g.
It is driven by predefined shapes and moves in a rectangular trajectory as shown in Fig.~\ref{fig:manipulation}.
We use the Gaussian function in (\ref{eqn:gs}) to generate the shapes, based on which the control messages are the coordinates of the center, the width and the amplitude of the function.
The control messages are broadcast to all modules in one CAN data frame at a rate of $60$\,Hz, so the refresh rate of shape is also $60$\,Hz, independent of system size.
The trajectory length is $20$\,cm and the average time for one cycle is about $2$\,s, resulting an average speed of $10$\,cm/s.
 
\section{DISCUSSION AND CONCLUSIONS}\label{sec:conclusion}
We present a control method for robotic surface that has system size-independent time delay.
Actuators are driven via broadcast communication and the control inputs are computed on-site, in contrast to previous methods that drive the actuators one-by-one or row-by-row.
We implement this control method in a $4\times 4$ pin array and experimentally confirm the system size-independent time delay.
The presence of actuator dynamics does not affect this scaling behavior.
Based on discrete cosine transform and matching pursuit algorithm, different shapes are efficiently approximated because less control messages are required comparing to standard sequential method.
We also show a compact way for generating shapes in object manipulation tasks, in which the center coordinates, width, and amplitude coefficients are directly sent to actuation modules.
Hence, our control method is more scalable and has the potential to control robotic surface with more actuators.

As a multi-actuator system, robotic surface benefits from a large number of actuators working together to accomplish various tasks, while suffers from the cost and complexity in configuring and controlling those many actuators.
Our proposed control method holds a great promise in reducing this burden and enable controlling a large number of actuators with easy.
Although the time delay scaling is only validated on a small set of actuation modules, and the control method is essentially open-loop, we demonstrate its scalable performance for large systems.
Comparing to sequential or row-column-based control method, it is worth noticing that the identifier of each module becomes useful information in computing the control inputs.
Ongoing works are focusing on theoretical controllability and closed-loop stability of this control method, as well as hardware and software upgrades for more compact designs and more demanding manipulation tasks.
Our method may be applicable to other multi-actuator systems in soft robotics, swarm robotics, and modular robotics.

\section*{ACKNOWLEDGMENT}
Y. Z. thanks Cheng Zhao, Shihua Li, Xin Xin, and Haifeng Xia for helpful discussions.



\begin{thebibliography}{99}


\bibitem{walker2017continuum}
I.~D. Walker, ``{Continuum robot surfaces: Smart saddles
  and seats},'' in \emph{{Mechatronics and {Robotics}
  {Engineering} for {Advanced} and {Intelligent} {Manufacturing}}}.\hskip 1em
  plus 0.5em minus 0.4em\relax Springer, Cham, 2017, pp. 97--105.

\bibitem{leithinger2014physical}
D.~Leithinger, S.~Follmer, A.~Olwal, and H.~Ishii, ``Physical telepresence:
  shape capture and display for embodied, computer-mediated remote
  collaboration,'' in \emph{Proceedings of the 27th annual {ACM} symposium on
  {User} interface software and technology}, ser. {UIST} '14.\hskip 1em plus
  0.5em minus 0.4em\relax New York, NY, USA: Association for Computing
  Machinery, 2014, pp. 461--470.

\bibitem{chen2023novel}
H.~Chen, W.~Tao, C.~Liu, Q.~Shen, Y.~Wu, L.~Ruan, and W.~Yang, ``A {Novel}
  {Refreshable} {Braille} {Display} {Based} on the {Layered} {Electromagnetic}
  {Driving} {Mechanism} of {Braille} {Dots},'' \emph{IEEE Transactions on
  Haptics}, vol.~16, no.~1, pp. 96--105, 2023.

\bibitem{nakagaki2016materiable}
K.~Nakagaki, L.~Vink, J.~Counts, D.~Windham, D.~Leithinger, S.~Follmer, and
  H.~Ishii, ``Materiable: {Rendering} {Dynamic} {Material} {Properties} in
  {Response} to {Direct} {Physical} {Touch} with {Shape} {Changing}
  {Interfaces},'' in \emph{Proceedings of the 2016 {CHI} {Conference} on
  {Human} {Factors} in {Computing} {Systems}}, ser. {CHI} '16.\hskip 1em plus
  0.5em minus 0.4em\relax New York, NY, USA: Association for Computing
  Machinery, 2016, pp. 2764--2772.

\bibitem{abtahi2018visuo-haptic}
P.~Abtahi and S.~Follmer, ``Visuo-{Haptic} {Illusions} for {Improving} the
  {Perceived} {Performance} of {Shape} {Displays},'' in \emph{Proceedings of
  the 2018 {CHI} {Conference} on {Human} {Factors} in {Computing} {Systems}},
  ser. {CHI} '18.\hskip 1em plus 0.5em minus 0.4em\relax New York, NY, USA:
  Association for Computing Machinery, 2018, pp. 1--13.

\bibitem{nakagaki2019inforce}
K.~Nakagaki, D.~Fitzgerald, Z.~J. Ma, L.~Vink, D.~Levine, and H.~Ishii,
  ``{inFORCE}: {Bi}-directional `{Force}' {Shape} {Display} for {Haptic}
  {Interaction},'' in \emph{Proceedings of the {Thirteenth} {International}
  {Conference} on {Tangible}, {Embedded}, and {Embodied} {Interaction}}, ser.
  {TEI} '19.\hskip 1em plus 0.5em minus 0.4em\relax New York, NY, USA:
  Association for Computing Machinery, 2019, pp. 615--623.

\bibitem{uriarte2019control}
C.~Uriarte, A.~Asphandiar, H.~Thamer, A.~Benggolo, and M.~Freitag,
  ``{Control strategies for small-scaled conveyor
  modules enabling highly flexible material flow systems},''
  \emph{{Procedia CIRP}}, vol.~79, pp. 433--438, 2019.

\bibitem{chen2024trajectory}
Z.~Chen, Z.~Deng, J.~S. Dhupia, M.~Stommel, and W.~Xu, ``Trajectory {Planning}
  and {Tracking} of {Multiple} {Objects} on a {Soft} {Robotic} {Table} {Using}
  a {Hierarchical} {Search} on {Time}-{Varying} {Potential} {Fields},''
  \emph{IEEE Transactions on Robotics}, vol.~40, pp. 351--363, 2024.

\bibitem{Wang2019design}
Y.~Wang, C.~Frazelle, R.~Sirohi, L.~Li, I.~D. Walker, and K.~E. Green, ``Design
  and {Characterization} of a {Novel} {Robotic} {Surface} for {Application} to
  {Compressed} {Physical} {Environments} *,'' in \emph{2019 {International}
  {Conference} on {Robotics} and {Automation} ({ICRA})}, 2019, pp. 102--108,
  iSSN: 2577-087X.

\bibitem{salerno2020ori-pixel}
M.~Salerno, J.~Paik, and S.~Mintchev, ``Ori-{Pixel}, a {Multi}-{DoFs} {Origami}
  {Pixel} for {Modular} {Reconfigurable} {Surfaces},'' \emph{IEEE Robotics and
  Automation Letters}, vol.~5, no.~4, pp. 6988--6995, 2020, conference Name:
  IEEE Robotics and Automation Letters.

\bibitem{tian2022soft}
Y.~Tian, G.~Fang, J.~S. Petrulis, A.~Weightman, and C.~C.~L. Wang, ``Soft
  {Robotic} {Mannequin}: {Design} and {Algorithm} for {Deformation}
  {Control},'' \emph{IEEE/ASME Transactions on Mechatronics}, pp. 1--10, 2022,
  conference Name: IEEE/ASME Transactions on Mechatronics.

\bibitem{adapa_adaptive_moulds2024adaptive}
A.~A. Moulds, ``Adaptive mould d100,'' 2024. [Online]. Available: https://adapamoulds.com/portfolio-item/adaptive-mould-d100/

\bibitem{smoot2019floor}
L.~S. Smoot, G.~D. Niemeyer, D.~L. Christensen, and R.~Bristow,
  ``{Floor system providing omnidirectional movement of
  a person walking in a virtual reality environment},'' US Patent
  US10\,416\,754B2, Sept., 2019.

\bibitem{johnson2023multifunctional}
B.~K. Johnson, M.~Naris, V.~Sundaram, A.~Volchko, K.~Ly, S.~K. Mitchell,
  E.~Acome, N.~Kellaris, C.~Keplinger, N.~Correll, J.~S. Humbert, and M.~E.
  Rentschler, ``{A multifunctional soft robotic shape
  display with high-speed actuation, sensing, and control},''
  \emph{{Nature Communications}}, vol.~14, no.~1, p.
  4516, July 2023, number: 1 Publisher: Nature Publishing Group.

\bibitem{robertson2019compact}
M.~A. Robertson, M.~Murakami, W.~Felt, and J.~Paik, ``A {Compact} {Modular}
  {Soft} {Surface} {With} {Reconfigurable} {Shape} and {Stiffness},''
  \emph{IEEE/ASME Transactions on Mechatronics}, vol.~24, no.~1, pp. 16--24,
  2019, conference Name: IEEE/ASME Transactions on Mechatronics.

\bibitem{liu2021robotic}
K.~Liu, F.~Hacker, and C.~Daraio, ``Robotic surfaces with reversible,
  spatiotemporal control for shape morphing and object manipulation,''
  \emph{Science Robotics}, vol.~6, no.~53, p. eabf5116, 2021, \_eprint:
  https://www.science.org/doi/pdf/10.1126/scirobotics.abf5116.

\bibitem{winck2013dimension}
R.~C. Winck and W.~J. Book, ``Dimension reduction in a feedback loop using the
  {SVD}: {Results} on controllability and stability,'' \emph{Automatica},
  vol.~49, no.~10, pp. 3084--3089, 2013.

\bibitem{follmer2013inform}
S.~Follmer, D.~Leithinger, A.~Olwal, A.~Hogge, and H.~Ishii, ``{inFORM}:
  dynamic physical affordances and constraints through shape and object
  actuation,'' in \emph{Proceedings of the 26th annual {ACM} symposium on
  {User} interface software and technology}, ser. {UIST} '13.\hskip 1em plus
  0.5em minus 0.4em\relax New York, NY, USA: Association for Computing
  Machinery, 2013, pp. 417--426.

\bibitem{xue2023arraybot}
Z.~Xue, H.~Zhang, J.~Cheng, Z.~He, Y.~Ju, C.~Lin, G.~Zhang, and H.~Xu,
  ``{ArrayBot}: {Reinforcement} {Learning} for {Generalizable} {Distributed}
  {Manipulation} through {Touch},'' June 2023, arXiv:2306.16857 [cs].

\bibitem{siu2018shapeshift}
A.~F. Siu, E.~J. Gonzalez, S.~Yuan, J.~B. Ginsberg, and S.~Follmer,
  ``{shapeShift}: {2D} {Spatial} {Manipulation} and {Self}-{Actuation} of
  {Tabletop} {Shape} {Displays} for {Tangible} and {Haptic} {Interaction},'' in
  \emph{Proceedings of the 2018 {CHI} {Conference} on {Human} {Factors} in
  {Computing} {Systems}}, ser. {CHI} '18.\hskip 1em plus 0.5em minus
  0.4em\relax New York, NY, USA: Association for Computing Machinery, 2018, pp.
  1--13.

\bibitem{stanley2016closed-loop}
A.~A. Stanley, K.~Hata, and A.~M. Okamura, ``Closed-loop shape control of a
  {Haptic} {Jamming} deformable surface,'' in \emph{2016 {IEEE} {International}
  {Conference} on {Robotics} and {Automation} ({ICRA})}, May 2016, pp.
  2718--2724.

\bibitem{leithinger2010relief}
D.~Leithinger and H.~Ishii, ``Relief: a scalable actuated shape display,'' in
  \emph{Proceedings of the fourth international conference on {Tangible},
  embedded, and embodied interaction}, ser. {TEI} '10.\hskip 1em plus 0.5em
  minus 0.4em\relax New York, NY, USA: Association for Computing Machinery,
  2010, pp. 221--222.

\bibitem{chen2011handbook}
J.~Chen, W.~Cranton, and M.~Fihn, \emph{Handbook of visual display
  technology}.\hskip 1em plus 0.5em minus 0.4em\relax Springer Publishing
  Company, Incorporated, 2011.

\bibitem{zhu2004practical}
H.~Zhu and W.~J. Book, ``{Practical {Structure} {Design}
  and {Control} for {Digital} {Clay}}.''\hskip 1em plus 0.5em minus 0.4em\relax
  American Society of Mechanical Engineers Digital Collection, 2004, pp.
  1051--1058.

\bibitem{zhu2006construction}
------, ``Construction and control of massive hydraulic
  miniature-actuator-sensor array,'' in \emph{2006 {IEEE} {Conference} on
  {Computer} {Aided} {Control} {System} {Design}, 2006 {IEEE} {International}
  {Conference} on {Control} {Applications}, 2006 {IEEE} {International}
  {Symposium} on {Intelligent} {Control}}, Oct. 2006, pp. 820--825, iSSN:
  2165-302X.

\bibitem{winck2012control}
R.~C. Winck, J.~Kim, W.~J. Book, and H.~Park, ``A control loop structure based
  on semi-nonnegative matrix factorization for input-coupled systems,'' in
  \emph{2012 {American} {Control} {Conference} ({ACC})}.\hskip 1em plus 0.5em
  minus 0.4em\relax IEEE, 2012, pp. 3484--3489.

\bibitem{winck2012control-1}
R.~C. Winck and W.~J. Book, ``{A {Control} {Loop}
  {Structure} {Based} on {Singular} {Value} {Decomposition} for
  {Input}-{Coupled} {Systems}}.''\hskip 1em plus 0.5em minus 0.4em\relax
  American Society of Mechanical Engineers Digital Collection, May 2012, pp.
  329--336.

\bibitem{winck2012command}
R.~C. Winck, J.~Kim, W.~J. Book, and H.~Park, ``{Command
  {Generation} {Techniques} for a {Pin} {Array} using the {SVD} and the
  {SNMF}},'' \emph{{IFAC Proceedings Volumes}}, vol.~45,
  no.~22, pp. 411--416, 2012.

\bibitem{winck2015svd}
R.~C. Winck and W.~J. Book, ``The {SVD} {System} for {First}-{Order} {Linear}
  {Systems},'' \emph{IEEE Transactions on Control Systems Technology}, vol.~23,
  no.~3, pp. 1213--1220, May 2015.

\bibitem{winck2016passivity}
------, ``Passivity and practical considerations for the {SNMF} {System},'' in
  \emph{2016 {IEEE} {International} {Conference} on {Robotics} and {Automation}
  ({ICRA})}.\hskip 1em plus 0.5em minus 0.4em\relax Stockholm, Sweden: IEEE,
  May 2016, pp. 2669--2674.

\bibitem{ferguson2020multiplicative}
K.~M. Ferguson, D.~Tong, and R.~C. Winck, ``Multiplicative valve to control
  many cylinders,'' in \emph{2020 {IEEE}/{ASME} {International} {Conference} on
  {Advanced} {Intelligent} {Mechatronics} ({AIM})}.\hskip 1em plus 0.5em minus
  0.4em\relax Boston, MA, USA: IEEE, July 2020, pp. 673--678.

\bibitem{jadhav2023scalable}
S.~Jadhav, P.~E. Glick, M.~Ishida, C.~Chan, I.~Adibnazari, J.~P. Schulze,
  N.~Gravish, and M.~T. Tolley, ``{Scalable {Fluidic}
  {Matrix} {Circuits} for {Controlling} {Large} {Arrays} of {Individually}
  {Addressable} {Actuators}},'' \emph{{Advanced
  Intelligent Systems}}, vol.~5, no.~8, p. 2300011, 2023.

\bibitem{Wang2023passively}
J.~Wang, M.~Sotzing, M.~Lee, and A.~Chortos, ``Passively addressed robotic
  morphing surface (parms) based on machine learning,'' \emph{Science
  Advances}, vol.~9, no.~29, p. eadg8019, 2023.

\bibitem{mallat1993matching}
S.~Mallat and {Zhifeng Zhang}, ``Matching pursuits with time-frequency
  dictionaries,'' \emph{IEEE Transactions on Signal Processing}, vol.~41,
  no.~12, pp. 3397--3415, Dec. 1993.

\bibitem{mallat2008wavelet}
S.~Mallat, \emph{{A {Wavelet} {Tour} of {Signal}
  {Processing}: {The} {Sparse} {Way},{Third} {Edition}}}, 3rd~ed.\hskip 1em
  plus 0.5em minus 0.4em\relax Elsevier Inc., Academic Press, Dec. 2008.

\bibitem{park1991universal}
J.~Park and I.~W. Sandberg, ``Universal approximation using
  radial-basis-function networks,'' \emph{Neural computation}, vol.~3, no.~2,
  pp. 246--257, 1991.

\end{thebibliography}
\end{document}